\documentclass[journal,twoside,web]{ieeecolor}
\usepackage{tmi}
\usepackage{cite}
\usepackage{amsmath,amssymb,amsfonts}
\usepackage{algorithmic}
\usepackage{graphicx}
\usepackage{textcomp}

\let\labelindent\relax
\usepackage{enumitem}
\usepackage{amsmath, amssymb,bm}
\usepackage{booktabs}
\usepackage{graphicx,epstopdf,subfigure}
\usepackage{multirow}
\usepackage{tabularx,threeparttable}
\usepackage{array}
\usepackage{comment}
\usepackage[hidelinks]{hyperref}
\usepackage{mathtools}
\usepackage{bbm}
\usepackage[ruled]{algorithm2e}

\def\BibTeX{{\rm B\kern-.05em{\sc i\kern-.025em b}\kern-.08em
    T\kern-.1667em\lower.7ex\hbox{E}\kern-.125emX}}
\begin{document}
\title{Temporal Memory Relation Network for Workflow Recognition from Surgical Video}
\author{Yueming Jin,~\IEEEmembership{Member,~IEEE}, Yonghao Long, Cheng Chen,~\IEEEmembership{Student Member,~IEEE}, \\ Zixu Zhao,~\IEEEmembership{Student Member,~IEEE}, Qi Dou,~\IEEEmembership{Member,~IEEE}, Pheng-Ann Heng,~\IEEEmembership{Senior Member,~IEEE}

\thanks{Manuscript received January 11, 2021; revised March 11, 2021; accepted March 21, 2021.
This work was supported by Key-Area Research and Development Program of Guangdong Province, China (2020B010165004), Hong Kong RGC TRS Project No.T42–409/18-R, National Natural Science Foundation of China with Project No. U1813204 and CUHK Shun Hing Institute of Advanced Engineering (project MMT-p5–20). (\emph{Corresponding author: Yueming Jin}.)}
\thanks{Y. Jin, Y. Long, C. Chen, and Z. Zhao are with the Department of Computer Science and Engineering, The Chinese University of Hong Kong, HKSAR, China. (e-mail: \{ymjin, yhlong, cchen, zxzhao\}@cse.cuhk.edu.hk).}
\thanks{Q. Dou is with the Department of Computer Science and Engineering, and T Stone Robotics Institute, The Chinese University of Hong Kong, HKSAR, China. (email: qdou@cse.cuhk.edu.hk).}
\thanks{P. A. Heng is with the Department of Computer Science and Engineering, The Chinese University of Hong Kong, HKSAR, China, and also with Guangdong-Hong Kong-Macao Joint Laboratory of Human-Machine Intelligence-Synergy Systems, Shenzhen Institutes of Advanced Technology, Chinese Academy of Sciences, China. (email: pheng@cse.cuhk.edu.hk).}}

\vspace{-6mm}
\maketitle
\begin{abstract}
Automatic surgical workflow recognition is a key component for developing context-aware computer-assisted systems in the operating theatre.
Previous works either jointly modeled the spatial features with short fixed-range temporal information, or separately learned visual and long temporal cues.
In this paper, we propose a novel end-to-end temporal memory relation network (TMRNet) for relating long-range and multi-scale temporal patterns to augment the present features.
We establish a long-range memory bank to serve as a memory cell storing the rich supportive information.
Through our designed temporal variation layer, the supportive cues are further enhanced by multi-scale temporal-only convolutions. 
To effectively incorporate the two types of cues without disturbing the joint learning of spatio-temporal features,
we introduce a non-local bank operator to attentively relate the past to the present. 
In this regard, our TMRNet enables the current feature to view the long-range temporal dependency, as well as tolerate complex temporal extents.
We have extensively validated our approach on two benchmark surgical video datasets, M2CAI challenge dataset and Cholec80 dataset.
Experimental results demonstrate the outstanding performance of our method, consistently exceeding the state-of-the-art methods by a large margin (e.g., 67.0\% v.s. 78.9\% Jaccard on Cholec80 dataset).

\end{abstract}


\section{Introduction}

\IEEEPARstart{C}{omputer} assisted surgery can help improve the quality of interventional healthcare, reduce intraoperative adverse events, and increase patient safety~\cite{maier2017surgical}.
Recognizing the surgical workflow is a fundamental enabler to design intelligent assistant systems in the operating room~\cite{padoy2019machine}.
Particularly, workflow recognition enables the systems to monitor and optimize surgical process, provide context-aware decision support, and generate early warning of potential deviations and anomalies~\cite{huaulme2020offline}.
The real-time analysis of workflow information can facilitate coordination and communication among the surgical team members~\cite{forestier2015automatic}.
\textcolor{black}{Another great usage is the automatic extraction of recorded database beyond intra-operative period,}
which can advance surgeon skill evaluation, surgical report documentation, surgeon educational training, and post-operative patient monitoring~\cite{padoy2012statistical,twinanda2017endonet,blum2010modeling}.
However, pure video-based workflow recognition is highly complex.
\textcolor{black}{Complicated surgical scenes lead to limited inter-phase variance but high intra-phase variance.}
Scenes are often blurred due to the frequent motion of the surgical instruments.
The changes in lighting conditions, inevitable visual occlusions from smoke and blood, and artifacts introduced by the lens cleaning process also increase the difficulty of surgery perception.

\begin{figure}[t]
	\centering
	\includegraphics[width=0.50\textwidth]{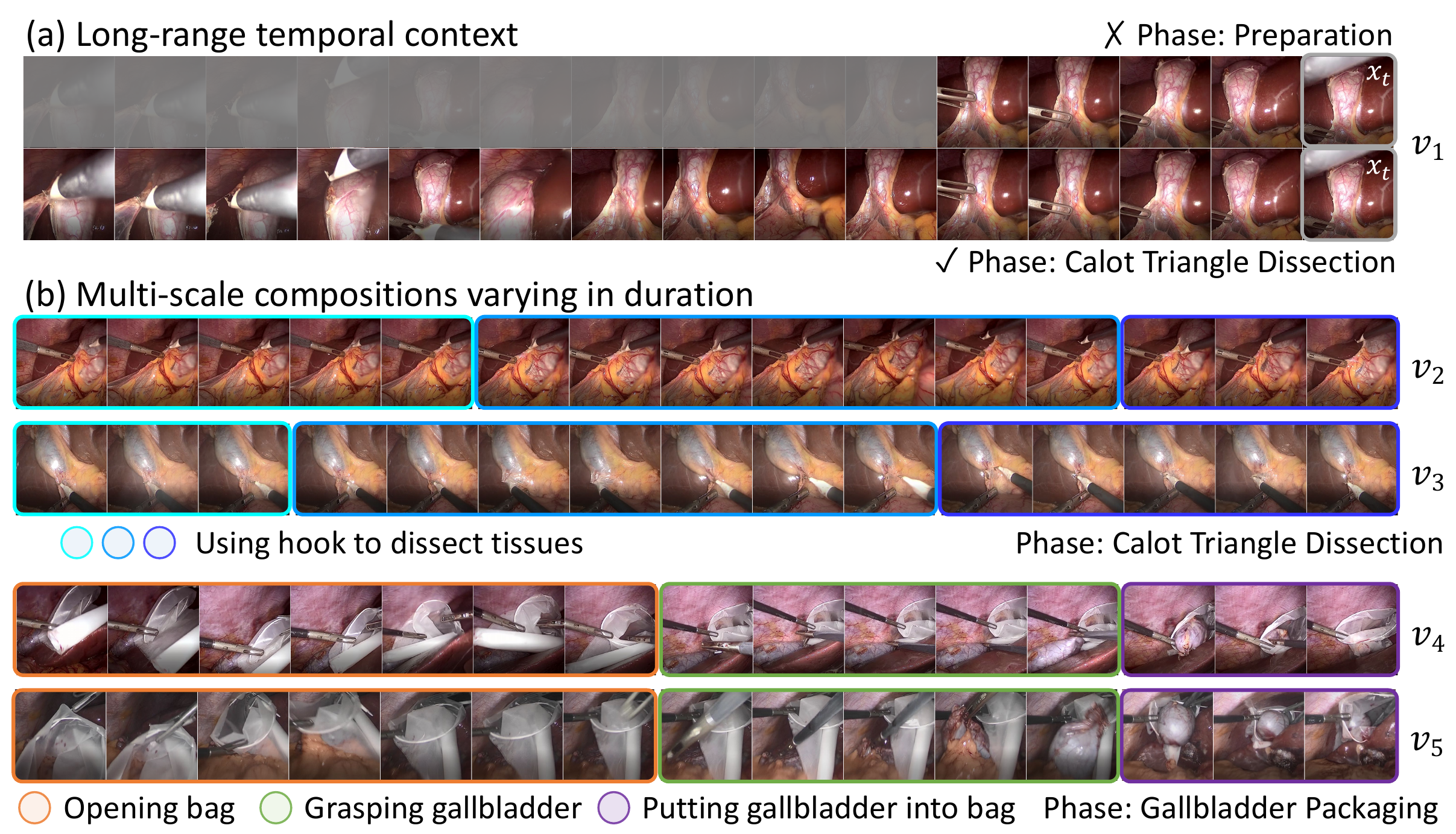}
	\vspace{-5mm}
	\caption{Illustration of two natural characteristics of surgical phases, taking the cholecystectomy procedure as an example.
	(a) Recognition is difficult without observing \emph{long-range temporal context}. The current frame $x_t$ in $v_1$ tends to be misclassified with a short-term view (top), instead of being recognized to ground truth phase with long-range cues (bottom). 
	(b) Each phase consists of \emph{multi-scale compositions varying in duration}, such as some homogeneous and repetitive actions ($v_2$ and $v_3$) or different and compositive motions ($v_4$ and $v_5$).
	For each case, 15 continuous frames with stride of 2 (about 30 seconds) are presented. Better zoom in for details.}
	\label{fig:intro}
	\vspace{-5mm}
\end{figure}

Capturing the long-range temporal dynamics effectively is of great importance for accurate workflow recognition, since surgical videos usually continue up to 30 minutes to 2 hours with each surgical phase lasting at least minute-level time.
To utilize the temporal information, one stream of recent works jointly learn the high-level spatio-temporal features using end-to-end recurrent convolutional networks (RCNets)~\cite{jin2018sv,jin2020multi} or 3D convolutional neural networks (CNNs)~\cite{funke2019using}.
However, those video models only collect information from a short-term view with 10s duration at most, limited by the GPU memory and computational resource.
Due to the large variety of surgical scenes in each phase and commonly existed artifacts, watching only short-term cues is not sufficient to achieve accurate recognition.
See Fig.~\ref{fig:intro}(a), recognizing phase of the current frame is difficult by watching 10s information.
Extending to longer-term context (about 30s) helps better perceive the present phase of Calot Triangle Dissection.
Another stream of solutions first extract features or prediction probabilities of isolated frames using CNN, and then input these pre-calculations into the temporal models for refinement, typically using hidden Markov models (HMMs) or recurrent neural networks (RNNs)~\cite{twinanda2017endonet,twinanda2018rsdnet,yengera2018less,miccai2019phase}.
Although covering up to the whole procedure, those separate training schemes hardly make the best use of the complementary information of visual and temporal features.
How to simultaneously take both advantages of end-to-end learning and long-range temporal information is crucial to achieving accurate workflow recognition.

We also identify another vital intrinsic property of surgical workflow.
As the surgical procedure can be described as a sequence of tasks at different granularity levels, including phases, steps, and activities, with gradually fine-grained definition~\cite{lalys2014surgical},
each phase can be naturally presented as the compositions of actions, in which we observe that these actions are characterized by high variability in their temporal duration.
For example, see Fig.~\ref{fig:intro} (b), the phase Calot Triangle Dissection can be broken down into the homogeneous and repetitive actions about using hook to dissect tissues, while varying in performing span.
The phase Gallbladder Packaging is composed of different actions, including opening bag, grasping gallbladder, and putting gallbladder into bag, also exhibiting large variations in temporal extents.
However, existing works scarcely take this nature into account.
They model the surgical video dynamics either using 3D convolutions with fixed kernel size~\cite{funke2019using}, or inputting the constant length of video clip representations into LSTM layers~\cite{twinanda2017endonet,zisimopoulos2018deepphase,miccai2019phase,jin2020multi}.
Both forms of variables are too rigid to capture the varieties of temporal space for complex surgical workflow analysis.

In this paper, we present a novel temporal memory relation network (TMRNet) for surgical workflow recognition, by relating long-range and multi-scale temporal supports for strengthening the features of current frame, with the end-to-end network training.
Concretely, we establish a long-range memory bank to store rich and time-indexed representations of the whole surgical procedure, serving as an auxiliary component to augment the standard video models.
A temporal variation layer is then proposed to enhance the long-range features from the memory bank with multi-scale convolutions in temporal space, which can reason about complex temporal patterns of surgical procedure.
Importantly, we design a non-local bank operator that enables our TMRNet to access such rich supportive context from the memory bank, without interrupting the end-to-end training process towards joint spatio-temporal feature learning.
Our method only accesses the previous information from memory bank when estimating each frame. Such online setup improves the applicability of our method for real-world surgery.
We extensively evaluate our method in online mode on two typical and publicly available surgical video datasets, including M2CAI challenge dataset and a large-scale Cholec80 dataset.
Our method outperforms existing state-of-the-art approaches significantly and consistently on both datasets.
Our main contributions are summarized as follows:

\begin{enumerate}
\item
We propose a novel framework, i.e., TMRNet, to accurately recognize workflow from surgical videos. Superior to previous short-term and rigid-scope temporal modeling, our end-to-end TMRNet can observe long-range temporal dependency and capture complex multi-scale temporal patterns for improving recognition accuracy.

\item
We build a long-range memory bank to provide the long-term supportive features encoding the past surgical frames.
The information is further augmented by our designed temporal variation layer with multi-scale temporal-only kernels, to tolerate large variations in duration of action components.

\item
We develop a non-local bank operator, to enhance the current spatio-temporal feature by attentively relating it to the information stored in the memory bank, also to remain the capability of end-to-end network training.

\item
We conduct extensive validations on two well-known and public benchmark surgical video datasets. Our method achieves a great performance gain, surpassing the state-of-the-art approaches by a large margin (e.g., $68.5\%$ v.s. $74.3\%$ Jaccard on M2CAI and $67.0\%$ v.s. $78.9\%$ Jaccard on Cholec80). \textcolor{black}{Code is available at \url{https://github.com/YuemingJin/TMRNet}.}

\end{enumerate}

\begin{figure*}[t]
	\centering
	\includegraphics[width=1\textwidth]{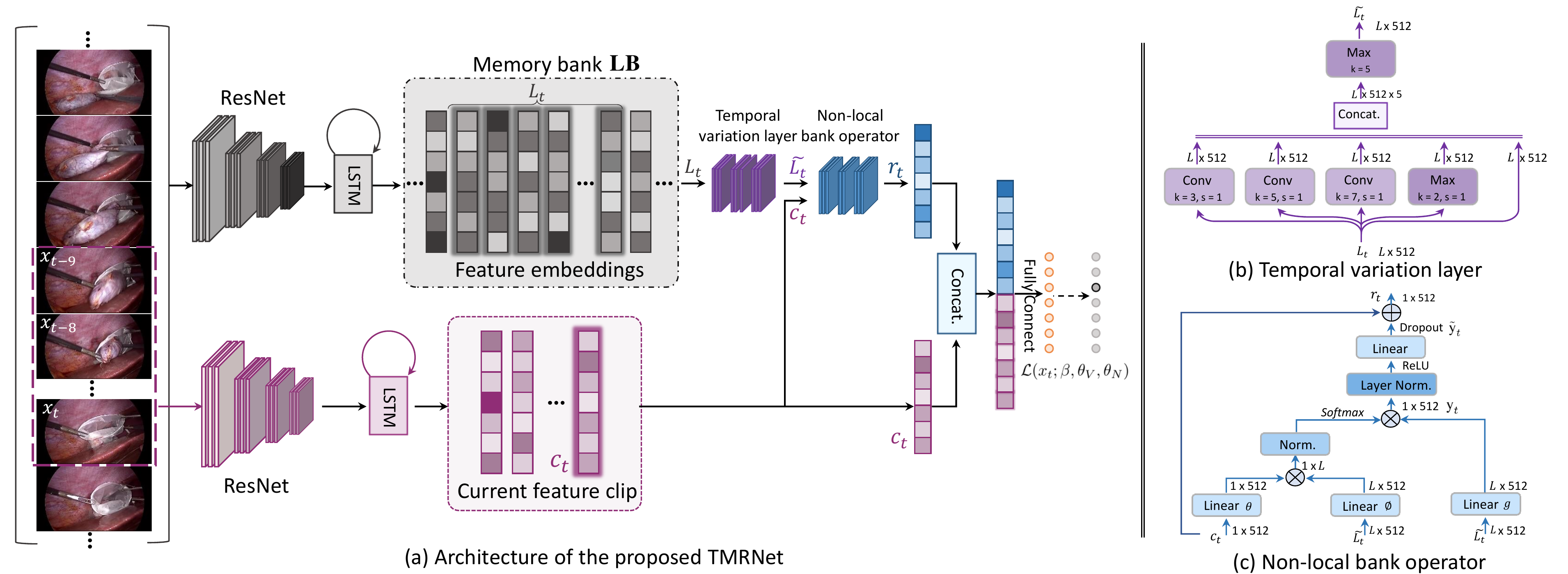}
	\vspace{-3mm}
	\caption{An overview of the proposed TMRNet for workflow recognition from surgical videos. We augment the conventional recurrent convolutional network (ResNet with LSTM) by a long-range memory bank. 
	\textcolor{black}{For frame $x_t$, we select long-range features $L_t$ from memory bank to augment current frame feature $c_t$, highlighting with the shadow.
	Temporal variation layer with multiple size kernels then tolerates the variations in temporal context of $L_t$.} Through the non-local bank operator, our model incorporates rich information to support current workflow prediction, from two essential perspectives for surgical videos, i.e., long-range temporal dynamics and multi-scale temporal extent.}
	\label{fig:overview}
	\vspace{-3mm}
\end{figure*}

\section{Related Work}

Extensive studies have been conducted for surgical workflow recognition, ranging from surgical phase recognition to fine-grained gesture, step and activity recognition. We aim at identifying the surgical phase in this work.
Visual data (i.e., surgical video) is the only ubiquitous data source, which can be routinely collected without disrupting the workflow in the minimally invasive surgery.
Therefore, the analysis based on pure visual information draws increasing attention to the community.
In this section, we also focus on reviewing this promising direction as it is highly related to our work.

Literature adopts surgical videos to conduct recognition tasks from two main aspects.
One stream targets at modelling high-level spatial and temporal features.
From the perspective of effective feature extraction, early studies utilize hand-crafted features to represent the spatial information, such as intensity value~\cite{zappella2013surgical}, gradient magnitude~\cite{blum2010modeling}, as well as shape, color and texture-based combinative descriptors~\cite{lalys2012framework}.
The performance is largely limited by the empirical design of these low-level features.
Meanwhile, previous works use linear statistical models to capture the temporal structure of surgical videos.
HMMs are primary choices with plentiful variations developed, such as left-right HMM~\cite{lalys2012framework}, Hidden semi-Markov Model~\cite{dergachyova2016automatic}, hierarchical HMM~\cite{twinanda2017endonet}.
Conditional Random Fields~\cite{tao2013surgical,quellec2014real,lea2015improved} and Dynamic Time Warping~\cite{blum2010modeling,lalys2013automatic} are also widely used for surgical temporal information modelling.
However, these methods rely on pre-defined dependence, which is insufficient to precisely represent some subtle yet essential motions with strong nonlinear dynamics.

With gratitude to high-level representation capability of deep learning, recent advances leverage CNNs and RNNs to extract spatial and temporal features, respectively.
Twinanda et al.~\cite{twinanda2017endonet} construct a 9-layer network based on AlexNet for visual feature extraction.
Jin et al.~\cite{jin2018sv} propose an end-to-end recurrent convolutional model called SV-RCNet. Deep residual network and LSTM module are seamlessly integrated into a unified framework to jointly capture spatio-temporal features of videos.
Lea et al.~\cite{lea2016segmental} design a segmental spatio-temporal CNN, comprised of a variation of VGG as spatial component and the large 1D convolutional filters as temporal component.
They then propose a temporal convolutional network called TCN, which is an encoder-decoder framework hierarchically capturing temporal relationships~\cite{lea2017temporal}.
Funke et al.~\cite{funke2019using} learn 3D CNNs to capture spatio-temporal cues from consecutive video frames.
Yi et al.~\cite{miccai2019phase} incorporate a three-step strategy into the CNN-LSTM network to alleviate the negative effect of hard samples. Data cleansing is first used to select hard samples and an online hard frame mapper is proposed to recognize the detected ones separately.
Current methods are either limited by short-term information capture, or are inferior due to the separate learning of spatio-temporal features.

Another important stream of works is dedicated to taking advantage of auxiliary information to boost workflow recognition performance, such as forming multi-task learning or multi-modal learning frameworks.
Several studies propose to simultaneously tackle the phase recognition and tool presence detection tasks, in order to exploit their close relatedness to complement the network training.
For example, Twinanda et al.~\cite{twinanda2017endonet} present a multi-task network, which consists of two branches sharing the early layers to extract the visual features. 
Zisimopoulos et al.~\cite{zisimopoulos2018deepphase} propose first to train a ResNet to recognize tool presence and then combine binary predictions and features to train an RNN for phase recognition. 
Jin et al.~\cite{jin2020multi} develop a complicated multi-task recurrent convolutional network based on SV-RCNet, with a deliberate design in each branch, i.e., CNN for tool recognition and RNN for phase recognition. A new correlation loss is proposed to provide constraints by minimizing the divergence of probability predictions from two branches. 10-second temporal information is considered in this framework.
In addition, aforementioned multi-tasking methods require the annotations of tool presence.
Apart from leveraging tool cues, Nakawala et al.~\cite{nakawala2019deep} present a Deep-Onto network which enhances deep models with knowledge management tools, ontology and production rules. By joint recognition of different workflow entities, semantic relations can be utilized to facilitate the recognition.
In addition, some works propose a multi-modal architecture to utilize other modalities such as optical flow~\cite{sarikaya2018joint} to enhance network training.
Recently, Qi et al.~\cite{qi2019deep} propose to leverage edge information extracted from raw frame as supplementary knowledge, and form a two-stream multi-modal network with raw data for inferring the surgical phase. 
These methods require extra annotation workload for multi-tasking, or introduce additional computation cost to calculate other modalities, which also hampers the increasing of input length.

\section{Methodology}

Fig.~\ref{fig:overview} (a) presents an overview of our temporal memory relation network for surgical workflow recognition.
A long-range memory bank is first introduced to store the long-term information, followed by the proposed temporal variation layer to tolerate the multi-scale motion cues.
By fully making use of the non-local bank operator, our TMRNet relates long-range and complex supports from the memory bank to the current features, and holds the capacity of end-to-end feature learning.

\subsection{RCNet based Long-range Memory Bank}
With recording the entire surgical procedure, most surgical videos present quite long duration and contain complex context.
Effectively relating the information that is distant in time to what is happening in the present can benefit a model to make accurate predictions on the current frame.
Our design leverages time information from a novel perspective compared with previous methods for surgical workflow recognition.
Rather than relying on the online-training modules (3D convolutional layer or LSTM) which only captures a short-term view, we propose to build an external long-range memory bank $\textbf{LB}$ to store the information of the entire video.
It intuitively acts as an auxiliary memory cell to aid recognition at the current time step.
There is no need to calculate gradients and do back-propagation for memory bank during the network training.
Thanks to few computational costs involved, very long-term past cues can be aggregated without disturbing the end-to-end learning process.

To build a memory bank for the surgical workflow recognition task, we propose to first utilize the unified recurrent convolutional network (SV-RCNet)~\cite{jin2018sv}, to compute the feature of each frame.
Considering that the duration of each phase in surgical videos is commonly longer than that of general natural video, 
instead of using a 3D CNN~\cite{wu2019long}, we leverage SV-RCNet to construct the memory bank.
The model is advanced by the seamless integration of a 50-layer residual network (ResNet) with an LSTM network, therefore has the capability of capturing longer cues with a few computational costs.

Specifically, to generate the feature $l_t$ of frame $x_t$ in the memory bank, we extract a video clip as the model input, which contains the current frame $x_t$ and a set of its previous frames as $\bm{x} \! = \! \{ x_{t-m}, \ldots, x_{t-1}, x_t \}$.
The model outputs a high-level spatio-temporal feature vector $l_t \in \mathbb{R}^d$ with $d \!=\! \text{512}$ for the current frame.
Clip-length temporal cues $[t-m, ..., t]$ therefore are embedded into the current frame feature $l_t$, which contributes to leverage the previous information to a large extent. 
We then compute the complete memory bank by passing the entire video using SV-RCNet model at regularly spaced intervals.
Formally, the long-range memory bank $\textbf{LB} \! = \! \{ l_{0}, l_{1}, \ldots, l_{T-1}\} $ forms a time-indexed feature list for a $T$-length video with time steps $[0, 1, \ldots , T-1]$.
It  provides information about when (time index) and which phase (feature) are being conducted in the whole surgical video, and can be efficiently computed in a single pass over the videos.
The stored time index is to help the model quickly locate the corresponding feature clip from memory bank when performing the augmentation for recognizing each frame.
Thanks to the external property of memory bank, various video models can be selected to generate the features in accordance with different task requirements.
Also, such design allows to include other supporting information, such as the tool usage feature and kinematic information.

\subsection{Modeling Variation with Multi-scale Temporal Kernels}

Each surgical phase is naturally composed of different fine-grained actions.
As shown in Fig.~\ref{fig:intro} (b), these actions demonstrate high variability in temporal duration.
Existing works model the temporal dynamics using fixed kernel size, which are too rigid to analyze the complex surgical workflow.
Tolerating the varieties in temporal extents is essential, especially for the complicated long-range temporal dependency from the memory bank, as it tends to contain more possibilities compared with the short-term information (see Fig.~\ref{fig:intro}).
Inception~\cite{szegedy2015going} proposed to account for different sizes of visual information in an image, by replacing the fixed-size 2D spatial kernels to multi-scale 2D spatial kernels.
Inspired by this work, we propose a temporal variation layer with multi-scale temporal kernels to enhance the representation capacity of long-range features.

As depicted in Fig.~\ref{fig:overview} (b), taking a long-range features $\bm{L_t} \in \mathbb{R}^{L\times512}$ with length step of $L$ as input, our temporal variation layer applies five temporal operations in total.
We perform various temporal convolutions in the first three operations, adopting $K$ kernels with different sizes $k = \{ 3, 5, 7 \} $.
A temporal max-pooling operation is then conducted with stride $s\!=\!1$, and kernel size $k\!=\!2$, to single out the max activation over local temporal region.
The final operation is a simple short connection to add the input feature $\bm{L_t}$ back for retaining the original information. 
All the operations maintain the number of temporal channels as input features at $L$.
Next, we consider how to fuse these five feature vectors which contain respectively distinctive temporal patterns.
Generalizing on recent architectures, we identify one layer design principle for CNNs:
the amount of layer parameters holds consistent for layer modularity.
The balance between the subspaces and their parameterization should also be retained.
A typical example is 2D CNN, in which expanding the spatial subspace is commonly associated with the reduction of semantic channel subspace.
Therefore, instead of directly concatenating the five outputs along the temporal channel axis,
we concatenate the outputs through expanding the number of channels by the factor of 5, and conclude the layer by a max-pooling with kernel size $k\!=\!5$ for channel reduction.
The final results then can revert to the original input dimension: $\tilde{\bm{L_t}} \in \mathbb{R}^{L\times512}$.

Accounting for the variations in action temporal span yields great benefits especially for long-term temporal dependency modeling.
We therefore inject our temporal variation layer before the bank operator, to process the long-range feature from the memory bank.
Intuitively, it augments the long-range feature by encoding various cues across time with multiple kernel sizes.
To this end, our model provides supports from two perspectives, i.e., aggregating long-range temporal context, and tolerating the temporal variations. 
We also consider incorporating various temporal convolutions into the feature bank operator, and discuss the results in our experiments.

\subsection{TMRNet with Non-local Bank Operator}
Ways to effectively leverage long-range and multi-scale past information are crucial.
We introduce a unified temporal memory relation network to incorporate temporal cues using a feature bank operator performed in a non-local scheme~\cite{wang2018non}. 

Formally, for recognizing the current frame $x_t$ in the online mode, we first form the long-range feature clip with length $L$ based on the time index $t$ from memory bank: $\bm{L_t} \! = \! \{ l_{t-L+1}, \ldots, l_t \}$, a subset of $\textbf{LB}$ that ended at the current time step with preceding $L-1$ frame features included.
$L$ has the largest number up to $t$, meaning that all the preceding features are considered.
The long-range feature clip $\bm{L_t}$ is further augmented by the temporal variation layer, outputting the refined representations $\tilde{\bm{L_t}}$ which also contain the motion compositional information.
Meanwhile, we employ SV-RCNet to extract the feature $c_t \in \mathbb{R}^{1 \times 512}$ of current frame $x_t$ by inputting the local short video clip $\bm{x} \! = \! \{ x_{t-n}, \ldots, x_{t-1}, x_t \}$.

The current feature $c_t$ and long-range multi-scale feature $\tilde{\bm{L_t}}$ are then forwarded to a feature bank operator.
Intuitively, the operator is using $c_t$ to attend $\tilde{\bm{L_t}}$, and adding the attended cues back to augment the current feature $c_t$.
We hereby design the operator based on the non-local block with the attention mechanism, where the non-local behavior is due to the fact that all information of $\tilde{\bm{L_t}}$ are considered in the operation.
Specifically, the original self-attention in the standard non-local block~\cite{wang2018non} is replaced with relating long-range feature $\tilde{\bm{L_t}}$ by $c_t$.
As illustrated in Fig.~\ref{fig:overview} (c), the soft weights are computed by first measuring similarities between $c_t$ and $\tilde{\bm{L_t}}$.
Here, we use an extension of Gaussian function to compute similarity in an embedding space:
\begin{equation}\label{equ2}
f( c_t, \tilde{\bm{L_t}}) = e^{{\theta(c_t)}^{T}\phi(\tilde{\bm{L_t}})},
\end{equation}
where $\theta(c_t) = W_{\theta} c_t$ and $\phi(\tilde{\bm{L_t}}) = W_{\phi}\tilde{\bm{L_t}}$ are two linear embeddings; the pairwise computation is conducted by matrix multiplication.
Then our non-local operation is defined as:
\begin{equation}\label{equ1}
\text{y}_{t} = \text{Softmax} (\frac{1}{\mathcal{N}} f( c_t, \tilde{\bm{L_t}})) g(\tilde{\bm{L_t}}),
\end{equation}
where $\mathcal{N}$ is the normalization factor for feature scaling~\cite{vaswani2017attention}, and the Softmax computation is used to perform the attentional behavior.
We also consider the unary function $g$ in the form of a linear embedding: $g(\tilde{\bm{L_t}}) = W_g\tilde{\bm{L_t}}$, computing the representations of $\tilde{\bm{L_t}}$.
Similar as the design principle of original non-local block~\cite{wang2018non}, we utilize operator $g$ to conduct the information transformation.
By linearly projecting the input signal to a latent embedding space, it avoids too heavy regularization with similarity matrix directly performed on the input signal.
To this end, the non-local operation acts as extracting the beneficial cues from $\tilde{\bm{L_t}}$ using the similarity score between $c_t$ and $\tilde{\bm{L_t}}$.
Additionally, as the training dataset of surgical video are commonly scarce, we integrate layer normalization~\cite{ba2016layer} and dropout strategy~\cite{srivastava2014dropout} to enhance the network regularization, which refine $\text{y}_{t}$ to $\tilde{\text{y}}_{t}$.
The improved attentional information $\tilde{\text{y}}_{t}$ is resummed with the current feature $c_t$ through a shortcut connection.
The non-local operator in this regard outputs a enhanced representation $r_t$ with original current feature complemented:
\begin{equation}\label{equ3}
r_{t} = \tilde{\text{y}}_{t} + c_{t}.
\end{equation}
We then concatenate the result $r_t$ of non-local bank operator with $c_t$, obtaining a 1024-dimensional feature vector. 
It is forwarded to two fully connected layers to generate the recognition predictions.
Note that the final feature vector includes current feature $c_t$ twice by first conducting element-wise summation via shortcut connection in non-local bank operator, and then feature concatenation. 
Both fusion manners are essential by involving important cues of current feature from different perspectives.
Element-wise summation can further enhance the attentional feature $\tilde{\text{y}}_{t}$ for implicit performance improvement, while concatenation alleviates breaking the respectively initial properties of two features, current feature $c_t$ therefore is directly exposed to the following classifier for explicit result enhancement.

\subsection{Objective Function and Training Details}
\subsubsection{Objective Function}
We employ the softmax cross-entropy function to calculate the loss of this multi-class recognition task:
\begin{equation}
\label{eq:overall}
\mathcal{L} (x_t ;\beta_R, \beta_L, \theta_V, \theta_N) = -  \log \hat{p}_t^{z=g_t}(x_{t-n:t},\bm{L_t} ),
\end{equation}
where $\hat{p}_t^z$ represents the predicted probability of frame $x_t$ belonging to the phase class $z$;
$g_t$ denotes the ground truth label of frame $x_t$; 
$\beta_R, \beta_L, \theta_V, \theta_N$ indicates the parameters of backbone network (i.e., ResNet and LSTM), temporal variation layer, and non-local bank operator respectively.
Given the local short clip $x_{t-n:t}$ and corresponding long-range feature $\bm{L_t}$, our approach regularizes the model to $g_t$, which can jointly optimize the weights of $\beta_R, \beta_L $ for spatio-temporal feature learning and also involves weights $\theta_V, \theta_N$ in the entire end-to-end training process.

\subsubsection{Training Procedure}
Joint training of the entire model including memory bank (see Fig.~\ref{fig:overview}) is not feasible, due to the computational complexity of back-propagating through the entire long-range memory bank $\textbf{LB}$. 
Instead, we treat $\textbf{LB}$ as a fixed component and obtain each feature offline without subsequent updates in this first stage.
Note that we also tried the alternating online update strategy for model optimization~\cite{le1986learning}, but no obvious improvement on results obtained.
Such external storage with a single computation pass also avoids repetitive and redundant costs for network training.
Specifically, we train SV-RCNet to obtain the memory bank $\textbf{LB}$ in the first stage.
Given that in SV-RCNet, the parameter scale of ResNet is much larger than that of LSTM layer,
we initialize the weights of ResNet with a pre-trained model on ImageNet~\cite{he2016deep}, and randomly initialize LSTM weights with Xavier normal initializer.
Next, we train our TMRNet by initializing the weights of SV-RCNet $[\beta_R, \beta_L]$ with trained model from the first stage, and randomly initialize the parameters of remaining parts $[\theta_V, \theta_N]$.
The entire parameters of TMRNet are jointly optimized towards Eq.~\ref{eq:overall}, without disturbing the end-to-end training process of spatio-temporal features. 
To adjust the learning rates, partial data from the training set are split as the validation data in the previous two learning stages.
In order to make the best usage of data, we further add the validation data back for network fine-tuning.
To this end, three-stage learning procedure is formed.
In the inference procedure, we sequentially create the input frame clip from each video in the form of a sliding window, with each time shifting one frame forward.
Memory bank keeps real-time updates by gradually archiving the frame features.
The long-range supportive features of previous frames therefore can be directly picked up from the memory bank to augment the current frame feature without pre-calculation.
In this regard, the memory bank can be handled in an online setup and retain the online prediction capacity of our model.

\subsubsection{Implementation Details}
Our framework is implemented based on PyTorch using 4 NVIDIA Titan Xp GPUs for training. 
The multiple GPU setting enables the batch size to reach $400$ frames.
To avoid bias in the network learning, we include various samples which are randomly sampled from the whole training dataset.
We empirically set the clip number up to 40 and each length can reach $10$ seconds, i.e., $n+1\!=\!m+1\!=\!10$.
The length of supporting temporal features from memory bank is set as $30$ seconds, i.e., $L=30$, to yield its maximal efficacy (see Table~\ref{tab:length80} and Table~\ref{tab:lengthm2cai}).
We train the model using synchronous stochastic gradient descent, with the momentum of 0.9 and weighted decay of $5e\!-\!4$.
For learning rate, we initialize ResNet as $5e\!-\!6$, LSTM as $5e\!-\!5$ in the first stage to obtain memory bank.
In the second stage, the initial learning rates are decreased to $5e\!-\!7$ for SV-RCNet, $5e\!-\!6$ for other parts, considering that we use a well pre-trained model from the first stage to initialize the model parameters.
We remain the initial learning rates as $5e\!-\!7$ for SV-RCNet, $5e\!-\!6$ for other parts in the third fine-tuning stage.
The learning rates are divided by a factor of 10 when the validation loss plateaus in the first two stages, and remain the same as initialization in the third fine-tuning stage.
We totally train the TMRNet 4000 iterations, taking around 3 hours for the entire training.
We use one GPU configuration in video inference.
For data preprocessing, we follow the previous works~\cite{twinanda2017endonet,jin2018sv} to downsample the original videos from 25\emph{fps} to 1\emph{fps} to enrich the temporal information for network training. 
We resize the frames from the original resolution of $1920 \! \times \! 1080$ and $854 \! \times \! 480$ into $250 \!  \times \!  250$ to save memory and reduce network parameters.
The data augmentations with $224\times224$ cropping, random mirroring and color jittering are performed to enlarge the training dataset.

\section{Experiments}

We validate the effectiveness of our workflow recognition approach on two public benchmark datasets of surgical videos,
with method comparison and extensive ablation analysis.
We implement two different network backbones of ResNet and ResNeSt to demonstrate the general efficacy of our method.

\vspace{-2mm}
\subsection{Datasets and Evaluation Metrics}

\textbf{M2CAI.} We employ a public challenge dataset from MICCAI 2016 Modeling and Monitoring of Computer Assisted Interventions Challenge, referred to as M2CAI~\cite{Cholec}.
The dataset consists of 41 videos recording the cholecystectomy procedures. 
Videos are acquired at 25fps and each frame has a resolution of $1920 \times 1080$. 
These videos are segmented into 8 phases by experienced surgeons. 
The detailed definition and time statistics of each phase can be found in~\cite{jin2018sv}. 
For fair comparison, we exactly follow the same evaluation process in the benchmark challenge and also reported in previous works~\cite{twinanda2016single,jin2018sv,miccai2019phase}, where the dataset is divided into 27 videos for training and 14 videos for testing.
In our network training, 7 videos from training set are split as validated data for adjusting learning rates in the first two stages.

\textbf{Cholec80.} We utilize a larger surgical dataset of Cholec80, which is publicly released by the same challenge organizers~\cite{twinanda2017endonet}.
Cholec80 includes more cholecystectomy procedures, i.e., 80 videos in total recorded with the same 25fps frequency.
The frames have the resolution of $1920 \times 1080$ or $854 \times 480$.
Compared with M2CAI, Cholec80 is fully annotated with $7$ defined phases by experienced physicians.
Specifically, the first two phases of cholecystectomy procedure are merged into one phase and others remain.
This dataset also contains tool annotations indicating the presence of 7 tools in an image.
We also exactly follow the same evaluation process reported in previous works for this dataset~\cite{twinanda2017endonet,jin2018sv,miccai2019phase}, i.e., splitting Cholec80 into two subsets of equal size, with the first $40$ videos as training set and the rest $40$ videos as testing set.
8 videos from training set are split as validated data for adjusting learning rates in the first two stages.

For evaluation, we employ four commonly-used metrics to quantitatively evaluate the performance,
which have also been utilized in previous surgical workflow recognition works~\cite{twinanda2017endonet,jin2018sv,miccai2019phase}.
These measurements include precision (PR), recall (RE), jaccard (JA) and accuracy (AC).
PR, RE and JA validate the results in phase-level.
We first compute PR, RE and JA of each phase followed by  $\mathrm{PR}=\frac{|\mathrm{GT} \cap \mathrm{P}|}{|\mathrm{P}|}, \mathrm{RE}=\frac{|\mathrm{GT} \cap \mathrm{P}|}{|\mathrm{GT}|}, \mathrm{JA}=\frac{|\mathrm{GT} \cap \mathrm{P}|}{|\mathrm{GT} \cup \mathrm{P}|}$,
where $\mathrm{GT}$ and $\mathrm{P}$ respectively represent the ground truth set and prediction set of one phase.
We then average these values over all the phases and obtain PR, RE and JA of the entire video.
The AC represents a video-wise evaluation, which is defined as the percentage of frames correctly classified into the ground truths in the entire video.
Note that we perform all our experiments in the online mode, where the future information (frames at time $t \textgreater t_0$) is not accessible when we estimate the frame at time $t_0$.

\vspace{-2mm}
\subsection{Comparison with State-of-the-arts}

\begin{table}[t]
	\centering
	\caption{Phase recognition results using different approaches \\ on M2CAI challenge dataset.}
	\vspace{-1mm}
	\label{tab:comparem2cai}
		\resizebox{0.48\textwidth}{!}{
			\begin{tabular}{l | cccc}
				\toprule
				Method                   & Accuracy (\%)     & Precision (\%)     & Recall (\%)    & Jaccard (\%) \\
				\hline
				PhaseNet~\cite{twinanda2016single}               & $79.5 \pm 12.1$   & -  &  -  & $64.1 \pm 10.3$ \\
				SV-RCNet~\cite{jin2018sv}                  & $81.7 \pm 8.1$  & $81.0 \pm 8.3$  & $81.6 \pm 7.2$ & $65.4 \pm 8.9$ \\
				OHFM~\cite{miccai2019phase}                      & $85.2 \pm 7.5$  & -  & -  & $68.8 \pm 10.5$  \\ \hline
				TMRNet (ResNet)       & $\pmb{86.3 \pm 9.8}$  & $\pmb{87.1 \pm 7.4}$  & $\pmb{87.8 \pm 6.1}$ & $\pmb{74.3 \pm 7.1}$ \\ 
				TMRNet (ResNeSt)     & $\pmb{87.0 \pm 8.6}$  & $\pmb{87.8 \pm 6.9}$  & $\pmb{88.4 \pm 5.3}$ & $\pmb{75.1 \pm 6.9}$ \\
				\bottomrule
			\end{tabular}}
			\\
			\vspace{-3mm}
\end{table}

\begin{table}[t]
	\centering
	\caption{Phase recognition results using different approaches \\ on Cholec80 dataset.}
	\vspace{-1mm}
	\label{tab:compare80}
		\resizebox{0.48\textwidth}{!}{
			\begin{tabular}{l | cccc}
				\toprule
				Method                   & Accuracy (\%)     & Precision (\%)     & Recall (\%)    & Jaccard (\%) \\
				\hline
				PhaseNet~\cite{twinanda2017endonet}                 & $78.8 \pm 4.7$  & $71.3 \pm 15.6$  & $76.6 \pm 16.6$  & -\\
				SV-RCNet~\cite{jin2018sv}                       & $85.3 \pm 7.3$  & $80.7 \pm 7.0$  & $83.5 \pm 7.5$ & - \\
				OHFM~\cite{miccai2019phase}                     & $87.3 \pm 5.7$  & -  & -  & $67.0 \pm 13.3$\\ \hline
				$\text{EndoNet}^{\ast}$~\cite{twinanda2017endonet}                  & $81.7 \pm 4.2$  & $73.7 \pm 16.1$  & $79.6 \pm 7.9$   & -\\
				$\text{EndoNet+LSTM}^{\ast}$~\cite{twinanda2017vision}                    & $88.6 \pm 9.6$  & $84.4 \pm 7.9$  & $84.7 \pm 7.9$  & -\\
				$\text{MTRCNet-CL}^{\ast}$~\cite{jin2020multi}  & $89.2 \pm 7.6$  & $86.9 \pm 4.3$  & $88.0 \pm 6.9$  & -\\ \hline
				TMRNet (ResNet)    & $\pmb{89.2 \pm 9.4}$  & $\pmb{89.7 \pm 3.5}$  & $\pmb{89.5 \pm 4.8}$ & $\pmb{78.9 \pm 5.8}$ \\
				TMRNet (ResNeSt)    & $\pmb{90.1 \pm 7.6}$  & $\pmb{90.3 \pm 3.3}$  & $\pmb{89.5 \pm 5.0}$ & $\pmb{79.1 \pm 5.7}$ \\
				\bottomrule
			\end{tabular}}
			\\
			\vspace{0.5mm}
			\scriptsize Note: the $\ast$ means methods of multi-task learning with extra tool labels needed. ~~~~~~~~~~~~~~~~~~~~
			\vspace{-3mm}
\end{table}

We compare our proposed method with several well-known or the state-of-the-art approaches for surgical workflow recognition on the two datasets.
PhaseNet extracts visual features by 9-layer CNN, followed by hierarchical HMM (HHMM) for temporal refinement on the prediction probabilities of whole videos (results quoted from~\cite{twinanda2016single,twinanda2017endonet}).
SV-RCNet \cite{jin2018sv} seamlessly integrates 50-layer ResNet and LSTM to jointly learn spatial and temporal features. 
OHFM \cite{miccai2019phase} also uses 50-layer ResNet to extract features and designs a three-step strategy to alleviate the negative effect of hard samples.
Predictions of all the previous frames from step-2 classifier are inputted to LSTM for mapping the current hard frame to its correct phase.
As Cholec80 dataset simultaneously provides tool presence annotations, the advanced methods on this dataset tend to leverage them to form a multi-tasking paradigm.
EndoNet~\cite{twinanda2017endonet} is a 9-layer multi-task CNN and HHMM is also employed for temporal regularization.
EndoNet+LSTM~\cite{twinanda2017vision} replaces HHMM by LSTM to enforce the sequential constraints on the visual feature from EndoNet.
MTRCNet-CL~\cite{jin2020multi} is a multi-task recurrent convolutional network, with a new correlation loss designed to minimize the prediction divergences from phase and tool branches.
We implement our TMRNet with 50-layer ResNet~\cite{he2016deep} as the backbone following the previous methods~\cite{miccai2019phase,jin2020multi} for a fair comparison.
Also, we use a more powerful state-of-the-art classification model (ResNeSt)~\cite{zhang2020resnest} to establish TMRNet, to investigate how accurate performance can be achieved towards real-world clinical applicability.

The results of different methods on M2CAI and Cholec80 datasets are presented in Table~\ref{tab:comparem2cai} and Table~\ref{tab:compare80}, respectively.
The standard deviations show the variation of results on different testing videos.
We see that all the other methods including ours outperform PhaseNet, indicating that deeper network with the non-linear temporal modeling of LSTM can extract more discriminative features for accurate workflow recognition.
With the same network backbone, OHFM although considers the long-range predictions up to the first frame for current frame recognition, its encoding process of spatial and temporal information is separated in different stages.
Our TMRNet attains much better performance on both datasets, demonstrating the importance of joint spatio-temporal feature learning for surgical workflow recognition.
Our TMRNet also outperforms the typical end-to-end training network SV-RCNet, by wisely incorporating longer temporal support.
To this end, compared with approaches only using phase label, our method boosts AC of phase recognition from $85.2\%$ to $86.3\%$ on M2CAI dataset, and from $87.3\%$ to $89.2\%$ on Cholec80 dataset.
We also achieve a large gain regarding JA on both datasets (over $5\%$ on M2CAI and $10\%$ on Cholec80).
Notably, the accuracy improving scopes of our TMRNet is larger on the Cholec80 dataset than M2CAI. The underlying reason could be that Cholec80 is a larger dataset with more complex and challenging issues, where effectively using long-range support and elaborately analyzing multi-scale compositions are important to recognize such challenging cases.

We further compare with the state-of-the-art multi-task learning methods on Cholec80 dataset in Table~\ref{tab:compare80} which requires the extra tool presence labels.
EndoNet and EndoNet+LSTM perform the separate visual and temporal feature learning.
Although they include additional tool information to do phase recognition, our TMRNet still achieves superior performance.
By leveraging high correlation between phase and tool recognition tasks, MTRCNet-CL also achieves high performance of $89.2\%$ in AC.
However, it utilizes limited temporal information of 10 seconds with the multi-GPU configuration.
Considering the longer-range temporal cues from memory bank, our TMRNet demonstrates its efficacy by the improvements on other two essential metrics, i.e., increasing PR from $86.9\%$ to $89.7\%$, RE from $88.0\%$ to $89.5\%$.
We also report the performance of our TMRNet with ResNeSt backbone, achieving the consistent improvement on both datasets.
It peaks AC performance over $90\%$ on Cholec80, narrowing gap from clinical usability.

\vspace{-2mm}
\subsection{Effectiveness of Key Components}

\begin{table}[t]
	\centering
	\caption{Recognition results on Cholec80 with two backbones \\ for analyzing different network components.}
	\vspace{-1mm}
	\label{tab:component}
		\resizebox{0.49\textwidth}{!}{
			\begin{tabular}{ l | l | c|cccc}
				\toprule
				\multicolumn{7}{c}{Backbone: ResNet}  \\ \hline
				 Method     &Cue    & Operator        & Accuracy (\%)     & Precision (\%)     & Recall (\%)    & Jaccard (\%) \\
				\hline
				 Baseline   & SR     &-          & $85.3 \pm 9.9$  & $82.4 \pm 5.4$  & $85.5 \pm 4.6$  & $70.4 \pm 7.2$  \\
				 TMRNet$^-$  & LR       & NL        & $88.6 \pm 9.8$  & $88.9 \pm 3.1$  & $89.1 \pm 3.4$  & $77.9 \pm 5.4$  \\
				 TMRNet' & LR+MS    & WAve    & $88.9 \pm 8.8$  & $89.6 \pm 3.5$  & $88.9 \pm 5.4$  & $78.2 \pm 6.0$  \\
				 TMRNet & LR+MS  & NL & $\pmb{89.2 \pm 9.4}$  & $\pmb{89.7 \pm 3.5}$  & $\pmb{89.5 \pm 4.8}$ & $\pmb{78.9 \pm 5.8}$ \\ \hline \hline
				 \multicolumn{7}{c}{Backbone: ResNeSt}  \\ \hline
				 Method     &Cue    & Operator        & Accuracy (\%)     & Precision (\%)     & Recall (\%)    & Jaccard (\%) \\
				\hline
				 Baseline      &SR      &-                  & $86.9 \pm 7.9$  & $83.9 \pm 5.1$  & $86.2 \pm 5.3$  & $72.0 \pm 8.2$  \\
				 TMRNet$^-$ & LR         & NL                 & $89.3 \pm 8.4$  & $89.6 \pm 3.6$  & $89.4 \pm 3.5$  & $78.5 \pm 5.2$  \\
				 TMRNet' & LR+MS    & WAve   & $88.9 \pm 8.4$  & $88.8 \pm 4.0$  & $88.6 \pm 5.8$  & $76.6 \pm 7.2$  \\
				 TMRNet & LR+MS &  NL & $\pmb{90.1 \pm 7.6}$  & $\pmb{90.3 \pm 3.3}$  & $\pmb{89.5 \pm 5.0}$ & $\pmb{79.1 \pm 5.7}$ \\
				\bottomrule
			\end{tabular}}
\end{table}

\begin{table}[t]
	\centering
	\caption{P-values for statistical analysis of all ablation settings for our proposed method.}
	\vspace{-1mm}
	\label{tab:pvalue}
		\resizebox{0.49\textwidth}{!}{
			\begin{tabular}{l |l | cccc}
				\toprule
				Backbone & Method             & Accuracy (\%)     & Precision (\%)     & Recall (\%)    & Jaccard (\%) \\
				\hline
				\multirow{3}{*}{ResNet}
				&Baseline~v.s.~TMRNet$^-$               & $2e^{-8}$   & $2e^{-8}$  &  $3e^{-7}$  & $5e^{-8}$ \\
				&TMRNet$^-$ \text{~v.s.~Ours}               & $0.004$  & $0.013$  & $0.093$ & $0.029$ \\
				&TMRNet'~v.s.~Ours          & $0.001$  & $0.368$ & $0.004$  & $0.021$  \\ \hline 
				\multirow{3}{*}{ResNeSt}
				&Baseline~v.s.~TMRNet$^-$                & $1e^{-4}$   & $1e^{-7}$  &  $5e^{-5}$  & $6e^{-6}$ \\
				&TMRNet$^-$\text{~v.s.~Ours}                & $0.028$  & $0.024$  & $0.032$ & $0.015$ \\
				&TMRNet'~v.s.~Ours           & $0.036$  & $0.091$ & $0.033$  & $0.007$  \\
				\bottomrule
			\end{tabular}}
\end{table}

We conduct ablation experiments to validate the effectiveness of different key components in the proposed method:
(1) Baseline: we train the pure backbone networks with short-range (SR) input frame of 10 seconds as the baseline of our experiments;
(2) $\text{TMRNet}^{-}$: we train our network with the memory bank providing long-range (LR) support features of 30 seconds, and use non-local (NL) operator to reference features from memory bank;
(3) $\text{TMRNet'}$: we employ long-range and multi-scale temporal information (LR+MS), and use the operator of a fully connected layer to weighted average (WAve) temporal channel of $\tilde{\bm{L_t}}$ to generate feature $r_t$;
(4) TMRNet: we employ long-range and multi-scale temporal information (LR+MS) and use non-local operator (NL) to reference features, i.e., our complete proposed model.
Two different network backbones, ResNet and ResNeSt, are employed to demonstrate the flexibility and general efficacy of our method.

\begin{figure}[t]
	\centering
	\includegraphics[width=0.48\textwidth]{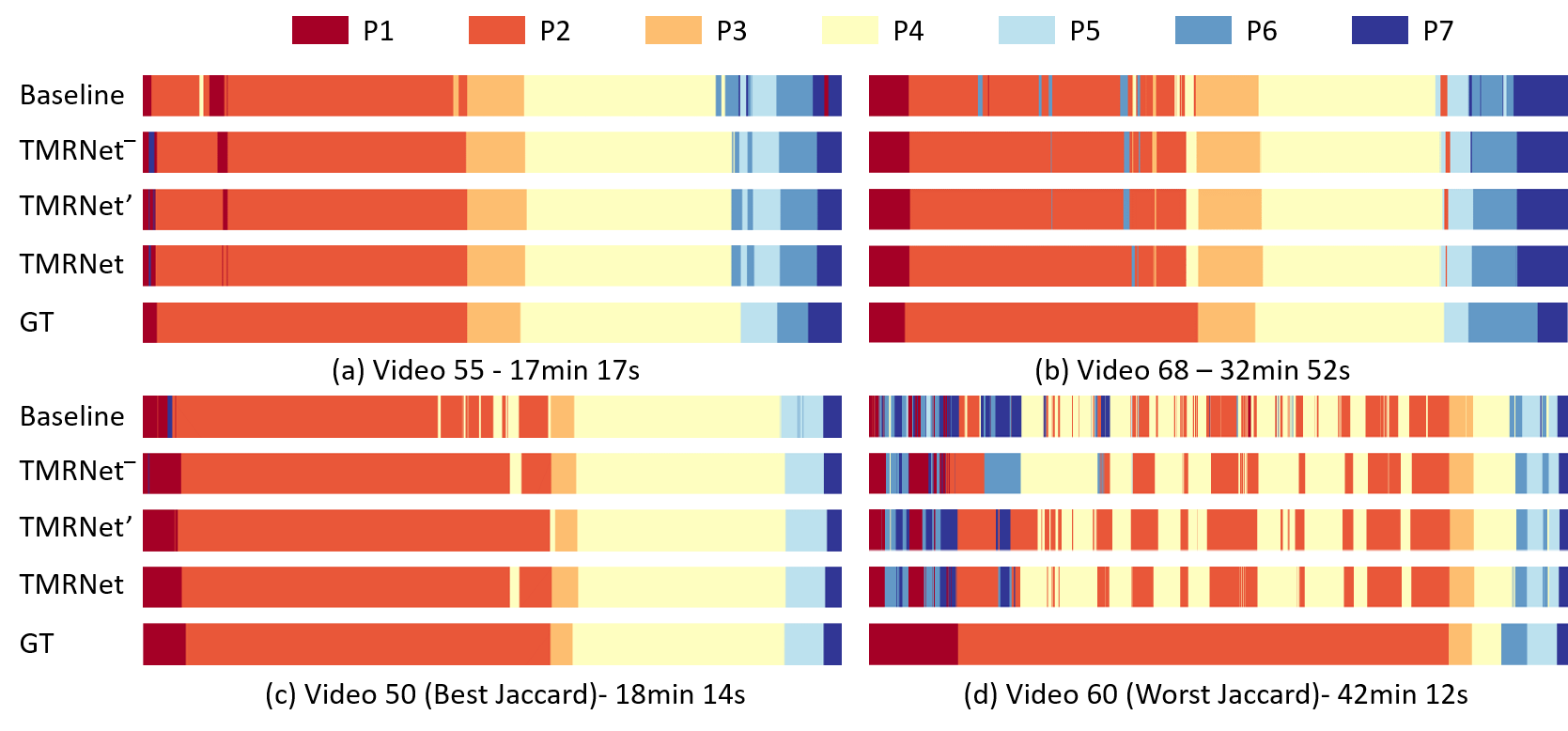}
	\vspace{-3mm}
	\caption{Color-coded ribbon illustration of seven phases (P1-P7) during four complete surgical videos, whose horizontal axis represents the time progression. In each case, from top to bottom are the results from our four ablation settings with ResNet backbone and the ground truth. We scale the temporal axes for better visualization.}
	\label{fig:result_phase}
\end{figure}

In Table~\ref{tab:component}, we observe that ResNet backbone network (Baseline) obtains reasonable recognition results with 85.3\% AC and 70.4\% JA.
On top of that, adding the long-range cues from our memory bank ($\text{TMRNet}^{-}$) steadily boosts the performance in all four evaluation metrics with 3\%-7\% gain, achieving 88.6\% AC and 77.9\% JA.
Comparing the results of $\text{TMRNet}^{-}$ and TMRNet, we can see that our multi-scale temporal kernels in temporal variation layer further improve the workflow recognition.
We then compare $\text{TMRNet'}$ and TMRNet, to validate the effectiveness of another key component of non-local operator to fuse features in temporal space. 
Note that with the aim of validating a complete non-local operator, in the configuration of $\text{TMRNet'}$, we weighted average $\tilde{\bm{L_t}}$ to generate the final output of the non-local operator, i.e., the feature $r_t$, rather than other intermediate representations.
We can see that compared with weighted average operation, employing non-local bank operator achieves the superior results.
Our full TMRNet with ResNet backbone finally obtains 89.2\% AC and 78.9\% JA.
Although the final improvement is attributed to both kinds of incorporated cues, long-range temporal information contributes much more than multi-scale patterns in accurate recognition.
Using another backbone of ResNeSt, we can also see the same consistent improvements by equipping the model with proposed components.
Although demonstrating a stronger baseline with ResNeSt, our TMRNet with long-range multi-scale temporal modeling still achieves stable improvements, peaking the performance at 90.1\% AC and 79.1\% JA.

To validate the stability of our method, we have repeated the series of experiments in ResNet backbone five time with random initialization.
The obtained average results are highly close to our reported results, with differences within 0.2\% for all the evaluation metrics. The variation of these five runs, i.e., the standard deviation, is also smaller than 0.4, demonstrating the high stability and reproducibility of our method.

\subsubsection{Statistical analysis on significance} We compute p-values using wilcoxon signed-rank test when comparing different ablation settings.
The numbers are given in Table~\ref{tab:pvalue} with two network backbones.
It is observed that we get $p < 0.05$ in all the evaluation metrics comparing Baseline with $\text{TMRNet}^-$, indicating a significant improvement of introducing long-range temporal cues.
We can also see $p < 0.05$ in most evaluation metrics in other two settings, verifying the significance of utilizing multi-scale information and non-local operation for accurate workflow recognition.

\begin{figure}[t]
	\centering
	\includegraphics[width=0.48\textwidth]{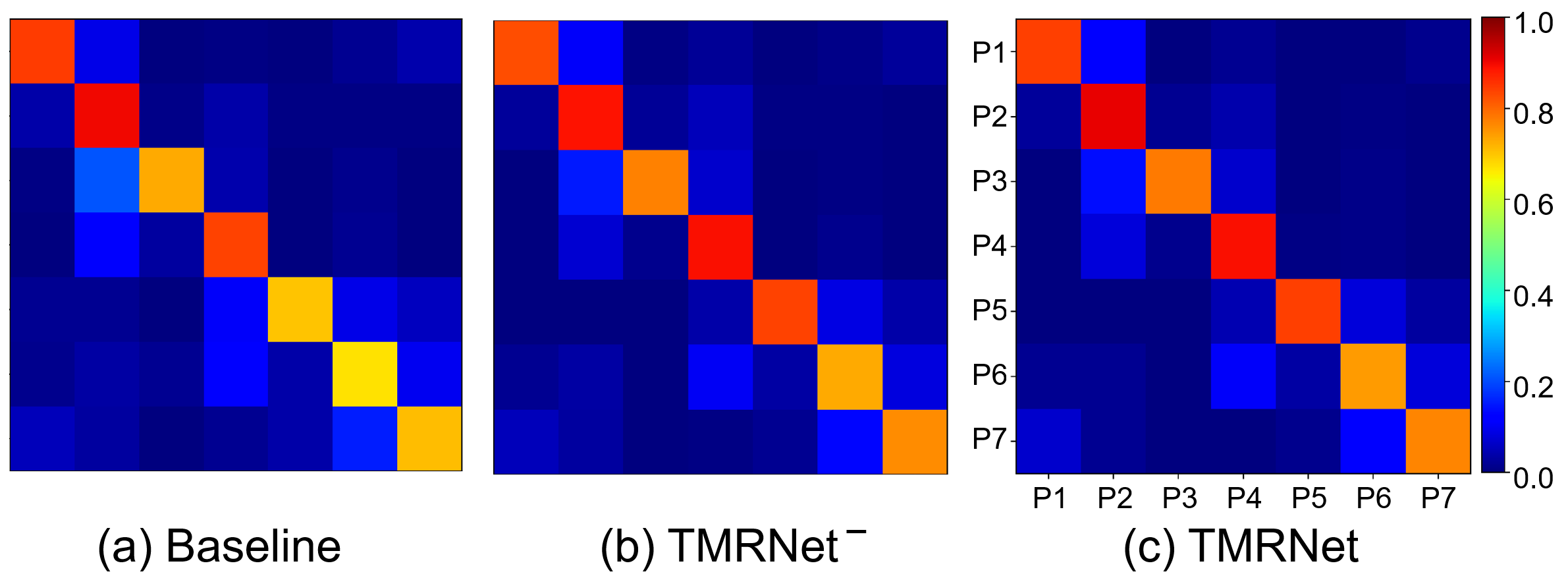}
	\caption{Confusion matrices visualized by color brightness of three methods
		(a) Baseline (SR), (b) $\text{TMRNet}^{-}$ (LR), and (c) TMRNet (LR+MS).
		In each matrix, X and Y-axis indicate predicted phase label and ground truth; 
		element (x, y) represents the empirical probability of predicting phase x given ground truth of phase y. Normalization is conducted along X-axis, with the summation of probabilities along X-axis equals to 1.}
	\label{fig:confusion_matrix}
	\vspace{-3mm}
\end{figure}

\subsubsection{Qualitative comparison} In Fig.~\ref{fig:result_phase}, we show the results of four settings on several complete surgical videos using color-coded ribbon.
It can be observed from (a) and (b) that, by gradually incorporating the long-term and multi-scale contexts, the model can better handle intermediate noisy patterns and hard frames of each surgical phase.
Our full model utilizing non-local operation achieves smoother results with more consistent phase predictions.
Such superiority is more obvious for the phase with longer duration (e.g. P2).
Along with the example in (c) when our TMRNet obtaining the best Jaccard, we can see that our method can accurately identify most phase transitions with deviation less than 10 seconds compared with ground truths. 
Our method is more precise to recognize transitions from P1 to P2, from P2 to P3, and from P5 to P6, with difference less than 5 seconds.
It plays a valuable role for computer- and robotic-assisted surgery to have a well preparation into next phase, such as automatically adjusting the configuration parameters in advance.
We also include results (d) when TMRNet obtaining the worst Jaccard.
Similar improvement trend is also shown especially in P1 and P2, but our method encounters obstacle to fully tackle the recognition especially for P2 in this challenging case.

We further present results on typical surgical actions in Fig.~\ref{fig:results}, where the prediction probabilities for frame $x_t$ under different ablation settings are indicated. 
It is observed that all key components can consistently enhance the model’s confidence towards correct predictions.

\subsubsection{Phase-level Comparison with Different Cues} 

Supportive temporal cues from memory bank play a core role in our method. 
In order to more comprehensively analyze the contribution of different supportive cues, we further visualize the confusion matrices to show the detailed results in phase-level.
The matrices with ResNet backbone are illustrated in Fig.~\ref{fig:confusion_matrix}.
We observe that from (a) to (c), the probability percentages on diagonals (recall) consistently increase while noisy misclassification patterns gradually decrease. Particularly, the incorrect recognition of P5 and P6 into P4, P7 into P1 is greatly alleviated by providing long-range multi-scale information.

\begin{figure*}[t]
	\centering
	\includegraphics[width=0.85\textwidth]{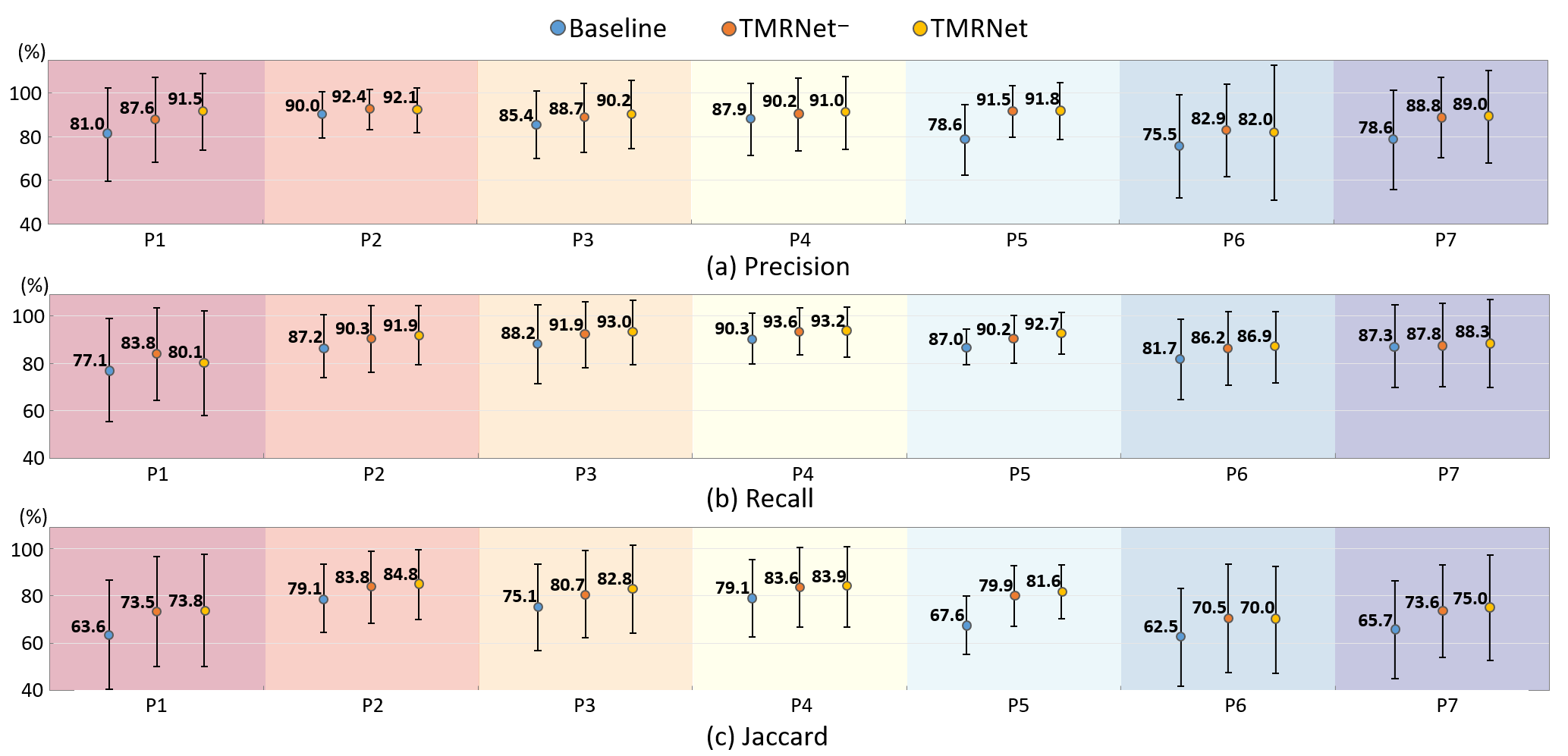}
	\vspace{-3mm}
	\caption{Phase-level results of (a) Precision, (b) Recall and (c) Jaccard of three settings:
		Baseline, $\text{TMRNet}^{-}$, and TMRNet. Average results and standard deviations of 40 test videos are shown by solid circles with numerical values and error bars in each chart.}
	\label{fig:bar}
	\vspace{-2mm}
\end{figure*}

\begin{table}[t]
	\centering
	\caption{Phase recognition results on Cholec80 dataset \\ with different length support.}
	\vspace{-1mm}
	\label{tab:length80}
		\resizebox{0.49\textwidth}{!}{
			\begin{tabular}{c|cccc}
				\toprule
				Length (sec.)      & Accuracy (\%)     & Precision (\%)     & Recall (\%)    & Jaccard (\%) \\
				\hline
				0                  & $85.3 \pm 9.9$  & $82.4 \pm 5.4$  & $85.5 \pm 4.6$  & $70.4 \pm 7.2$  \\
				10                  & $86.8 \pm 9.7$  & $86.2 \pm 4.2$  & $87.0 \pm 4.7$  & $74.2 \pm 6.2$ \\
				20                  & $88.3 \pm 9.3$  & $88.7 \pm 3.6$  & $88.0 \pm 5.1$  & $76.9 \pm 6.2$\\
				\pmb{30}            & $\pmb{88.6 \pm 9.8}$  & $88.9 \pm 3.1$  & $\pmb{89.1 \pm 3.4}$  & $\pmb{77.9 \pm 5.4}$ \\
				40                  & $\pmb{88.6 \pm 5.7}$  & $\pmb{89.1 \pm 4.0}$  & $88.9 \pm 6.1$  & $77.8 \pm 6.1$\\
				\bottomrule
			\end{tabular}}
\end{table}

\begin{table}[t]
	\centering
	\caption{Phase recognition results on M2CAI dataset \\ with different length support.}
	\vspace{-1mm}
	\label{tab:lengthm2cai}
		\resizebox{0.49\textwidth}{!}{
			\begin{tabular}{c | cccc}
				\toprule
				Length (sec.)      & Accuracy (\%)     & Precision (\%)     & Recall (\%)    & Jaccard (\%) \\
				\hline
				0             & $82.9 \pm 10.4$  & $80.8 \pm 11.1$  & $83.8 \pm 7.1$  & $67.2 \pm 9.5$ \\
				10                  & $83.8 \pm 11.1$  & $84.0 \pm 10.1$  & $84.6 \pm 8.1$  & $70.1 \pm 9.6$ \\
				20                  & $85.0 \pm 10.4$  & $85.2 \pm 9.2$  & $86.1 \pm 6.4$  & $71.5 \pm 8.8$\\
				\pmb{30}            & $\pmb{85.8 \pm 10.3}$  & $\pmb{86.2 \pm 8.3}$  & $\pmb{87.1 \pm 6.5}$  & $\pmb{73.2 \pm 7.8}$ \\
				40                  & $85.6 \pm 10.8$  & $85.8 \pm 6.6$  & $86.1 \pm 7.9$  & $72.8 \pm 7.0$\\
				\bottomrule
			\end{tabular}}
			\vspace{-6mm}
\end{table}

We further draw charts (see Fig.~\ref{fig:bar}) to provide more details by illustrating the numerical results of PR, RE and JA in each phase-level.
By gradually introducing long-range and multi-scale cues, our TMRNet achieves consistent improvements in most phases, across all the three metrics.
We observe our scheme shows large improvement of PR results especially in P1 and P3. 
Similarly, RE performances in P2, P5 and P6 have significant increase.
For JA, TMRNet dominates others across almost all the seven phases.

\vspace{-2mm}
\subsection{Importance of Long-range Information}
\subsubsection{Different lengths of supporting features}

Seeing what kinds of degree about temporal context is a key factor to recognize phase for each frame.
We analyze the impact of different lengths of supporting features from our long-range memory bank.
Specifically, we increase the window size $L$ of supporting features $\bm{L_t}$ with range $L$ $ \in [0,40]$ at a step of 10 seconds, while the length of frames input to network to obtain the features $c_t$ is kept as 10 seconds.
With $L$ $=0$, the network is trained without memory bank.
Multi-scale kernels are not incorporated in these settings for a clear and direct comparison (i.e., $\text{TMRNet}^{-}$ method).
Experiments are conducted on both Cholec80 and M2CAI datasets with ResNet backbone.
Results are reported in Table~\ref{tab:length80} and Table~\ref{tab:lengthm2cai}.

We observe that on Cholec80 dataset, gradual improvements are gained by increasing the temporal support length.
In particular, the metric AC improves from $85.3\%$ to $88.6\%$ when the length increases from $0$ seconds to $40$ seconds,
demonstrating the importance of integrating long-term temporal dependencies for phase recognition task.
Notably, as we build our memory bank upon the recurrent convolutional network, 10-second supporting features actually cover longer temporal cues, achieving better results than the pure ResNet with 10-second frame input.
Nevertheless, we also notice that the performance growth rate tends to be slower as the increase of supporting length.
The results even slightly decrease about $0.2\%$ in RE and $0.1\%$ in JA, when the length increases from 30 seconds to 40 seconds.
One of the underlying reasons might be that excessive long-range information may introduce too much irrelevant noise disturbing accurate recognition, such as artefacts from smoke and blood, and incompatible appearances due to large variation within one surgical phase.
The results on M2CAI dataset in Table~\ref{tab:lengthm2cai} also show a similar trend. 
To take maximal advantage of the memory bank, we employ 30-second features as long-range supports for our TMRNet.

\begin{figure*}[t]
	\centering
	\includegraphics[width=1\textwidth]{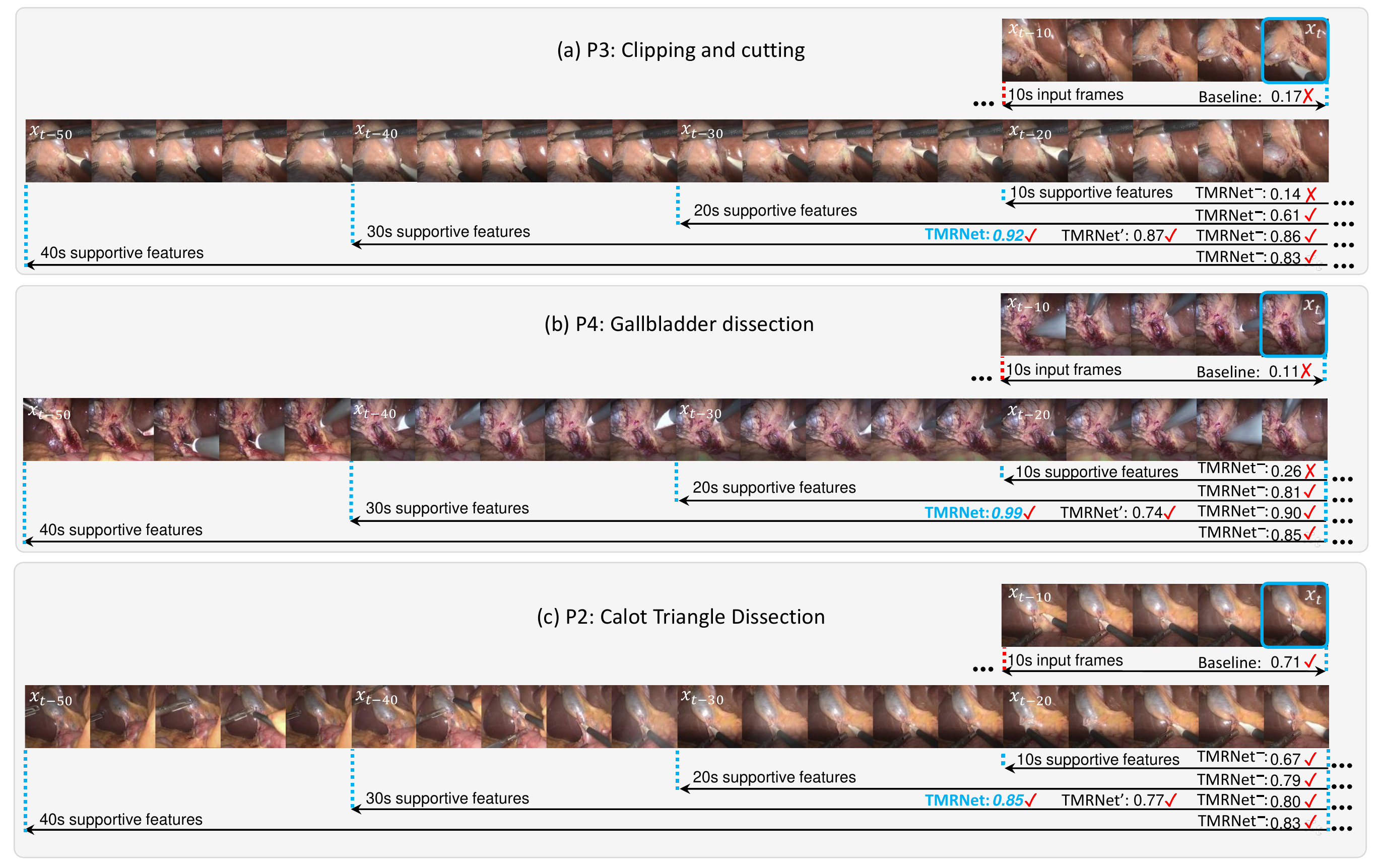}
	\vspace{-6mm}
	\caption{Visual comparison predicted by our methods under different ablation settings, and by $\text{TMRNet}^{-}$ using different-length supportive features $\bm{L_t}$ with window sizes ranging from 10s to 40s (covering 20s to 50s preceding frames thanks to using LSTM to build memory bank). Actions in three different phases during the cholecystectomy procedure are illustrated. For each phase, we present continuous frames with the stride of 2 to show the different-length supportive information. We list the ground-truth, and the prediction scores towards ground-truth phases for current frame $x_t$. 
	}
	\label{fig:results}
	\vspace{-3mm}
\end{figure*}

We further qualitatively illustrate a few examples showing the impact of the supporting length from memory bank in Fig.~\ref{fig:results}.
We compare the results on three different phases predicted by our network with the supportive feature length ranging from 10 seconds to 40 seconds.
Note that they actually can cover previous 20 second to 50 seconds information, thanks to using LSTM of 10-second temporal window to generate the supportive features.
Multi-scale kernels are also not incorporated in these settings for a clear and direct comparison (i.e., $\text{TMRNet}^{-}$ method).
We observe that the model fails to identify phase correctly with short-term view.
After seeing long-range extents, such as (a) more complete operation scenes or (b,c) typical actions, our model can boost the prediction results with higher probabilities towards ground truth phases.

\begin{table}[t]
	\centering
	\caption{Results on Cholec80 dataset using pure baseline \\ SV-RCNet with input clip in different lengths.}
	\vspace{-1mm}
	\label{tab:lengthLSTM}
		\resizebox{0.49\textwidth}{!}{
			\begin{tabular}{c|cccc}
				\toprule
				Length (sec.)      & Accuracy (\%)     & Precision (\%)     & Recall (\%)    & Jaccard (\%) \\
				\hline
				10                  & $\pmb{85.3 \pm 9.9}$  & $82.4 \pm 5.4$  & $85.5 \pm 4.6$  & $70.4 \pm 7.2$ \\
				\pmb{20}                  & $85.2 \pm 10.3$  & $\pmb{82.9 \pm 10.0}$  & $\pmb{87.2 \pm 3.9}$  & $\pmb{71.9 \pm 8.2}$\\
				30            & $83.5 \pm 11.2$  & $82.1 \pm 9.9$  & $85.9 \pm 8.9$  & $69.3 \pm 8.5$ \\
				40                  & $83.5 \pm 11.2$  & $82.3 \pm 8.1$  & $84.2 \pm 9.2$  & $69.0 \pm 9.7$\\
				\bottomrule
			\end{tabular}}
			\vspace{-2mm}
\end{table}

\subsubsection{Different manners to increase temporal cues}
Apart from using the long-range information from memory bank, another alternative way to increase temporal cues is directly increasing the temporal window of each input clip to LSTM layer in the pure baseline SV-RCNet.
Remaining the diversity of input clips by keeping the large number of video clips in each training batch is essential for network training.
We therefore use gradient accumulation strategy to enlarge the training batch while avoid introducing too much computational cost.
Specifically, we increase the temporal window of input clip to 20, 30, and 40 seconds to obtain features $c_t$, without using memory bank for training.
The clip number in each training batch keeps 40, which is the same as our proposed TMRNet.
For fair comparison, we found the optimal values of two vital parameters for model training, i.e., initialized learning rate and training iteration, and report the best results of different settings.
Generally, the role of initialized learning rate is consistent, with initializing ResNet as $5e\!-\!6$, LSTM as $5e\!-\!5$ yielding the best results. 
Regarding the training iteration, the models with longer input clips commonly require longer time for training.
Results with ResNet backbone on Cholec80 dataset are listed in Table~\ref{tab:lengthLSTM}.

We can see that compared to counterparts using supportive features in the same time length from memory bank (see Table~\ref{tab:length80}), directly increasing the length of input into LSTM fails to attain large improvements. Additionally, we can observe that performances even degrade when the input length increases to 30 and 40 seconds. One underlying reason of this phenomenon is that hidden state in LSTM layer hardly encodes and preserves quite long-range temporal cues. The long-range information also brings difficulties in LSTM training.

\begin{figure}[t]
	\centering
	\includegraphics[width=0.48\textwidth]{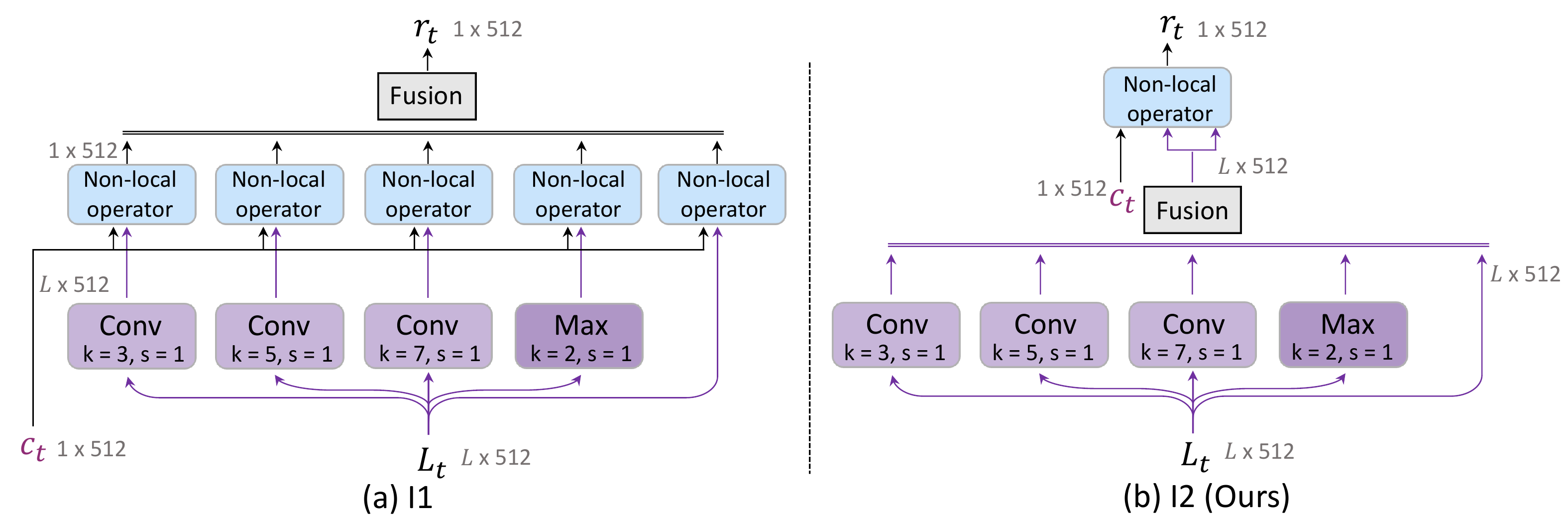}
	\vspace{-3mm}
	\caption{Two incorporating fashions of multi-scale temporal information.}
	\label{fig:incor}
	\vspace{-4mm}
\end{figure}

\vspace{-2mm}
\subsection{Detailed Analysis of Multi-scale Convolutions}
We explore the impact of the way to incorporate multi-scale temporal information, with different incorporating fashions (I1 or I2 in Fig.~\ref{fig:incor}), kernel types (fixed or multiple) and feature fusion styles (average or max).
The results are listed in Table~\ref{tab:multi}.
We see that incorporating multi-scale information in I2 fashion generally outperforms that in I1 way, with around $0.1\%$ improvement.
As shown in Fig.~\ref{fig:incor}, I1 performs different scale temporal convolutions on the long-range features with several times. The current feature is augmented with multiple non-local bank operators, and then fused to produce final predictions.
While I2 directly enhances the long-range features with multi-scale temporal kernels, then the non-local operator processes the enhanced long-range features with the current one.
Although both fashions can relate multi-scale temporal cues to the present feature with the non-local attention scheme, 
the late fusion of different non-local perfectives in I1 may break the learned attended structures, leading to less optimal performance.
We also notice that in the fixed-size kernel settings, smaller kernel size achieves larger contributions for temporal modeling, which is reasonable given that typical actions in cholecystectomy are generally frequent with each action lasting short duration.
The features encoded by the fixed kernel size of 3 or 5, even show stronger capacity compared with multi-scale features fused by average pooling operation.
By using max-pooling to single out the max activation over local temporal region, our multi-scale kernel yields its best benefits on tolerating complex actions in surgical procedure, particularly improving RE and JA performances.

\begin{table}[t]
	\centering
	\caption{Phase recognition results on Cholec80 dataset \\ for ablation study in multi-scale convolutions.}
	\vspace{-1mm}
	\label{tab:multi}
		\resizebox{0.48\textwidth}{!}{
			\begin{tabular}{c|c|c|cccc}
				\toprule
				Incor. & Kernel Size  &  Fusion           & Accuracy (\%)     & Precision (\%)     & Recall (\%)    & Jaccard (\%) \\
				\hline
				\multirow{3}{*}{I1}
				&$3\&5\&7$  & Ave         & $88.4 \pm 9.2$  & $\pmb{89.7 \pm 5.3}$  & $86.7 \pm 7.4$  & $75.8 \pm 8.3$ \\
				&$3\&5\&7$  & Max           & $88.3 \pm 9.3$  & $89.3 \pm 4.6$  & $87.3 \pm 6.0$  & $75.9 \pm 7.5$\\ \hline
				\multirow{5}{*}{\pmb{I2}}
				&3        & -         & $\pmb{89.2 \pm 9.3}$  & $\pmb{89.7 \pm 3.7}$  & $88.7 \pm 6.0$  & $78.3 \pm 6.4$ \\
				&5        & -           & $89.0 \pm 9.6$  & $89.6 \pm 3.4$  & $88.7 \pm 5.4$  & $78.4 \pm 6.1$\\
				&7        & -          & $88.7 \pm 9.5$  & $89.7 \pm 4.1$  & $88.5 \pm 6.1$  & $77.9 \pm 6.1$ \\ \cline{2-7}
				&$3\&5\&7$    &     Ave         & $88.8 \pm 9.1$  & $89.2 \pm 4.1$  & $88.9 \pm 5.3$  & $77.9 \pm 6.1$\\
				& \pmb{$3\&5\&7$}    &     \pmb{Max}         & $\pmb{89.2 \pm 9.4}$  & $\pmb{89.7 \pm 3.5}$  & $\pmb{89.5 \pm 4.8}$  & $\pmb{78.9 \pm 5.8}$\\
				\bottomrule
			\end{tabular}}
			\vspace{-2mm}
\end{table}

\section{Discussion}

Automatic surgical workflow recognition plays an indispensable role in advancing context-awareness in modern intelligent operating rooms.
Each phase in the surgical workflow is typically featured as a quite long-range and compositional procedure, which can be broken down into a set of subsector actions achieved during the corresponding phase.
Existing approaches broadly used LSTM network and 3D CNN to analyze the temporal space of surgical video, leading to less optimal performance regarding only short-term fixed-range temporal dependency modeling.
We present a novel temporal memory relation network (TMRNet) for surgical workflow recognition by relating the present feature with long-range multi-scale temporal context.
\textcolor{black}{Long-range memory bank and temporal variation layer are developed to provide such complex temporal context.}
Consistent performance improvements with two well-known backbones on public datasets demonstrate the effectiveness and architecture-independence of our scheme, which can also 
be easily integrated into various existing 2D and 3D CNN video models.

Way to capture temporal dependency of videos is a cornerstone and longstanding problem for accurate workflow recognition.
Previous works either employ recurrent convolutional networks to leverage the complementary spatio-temporal representations, in which input length is heavily limited by the computational resource (10 seconds in the state-of-the-art method~\cite{jin2020multi}).
Others form a two-stage framework, i.e., separately extracting the visual features at first, refined by the following HMM or LSTM for temporal regularization.
This stream of works prohibits the joint learning of visual and motion information, leading to a suboptimal solution for workflow recognition.
Ideally, increasing the computational resources in the former way to enlarge the input length and training batch size can promote the performance,
however, is hardly applicable in the real-world clinical circumstance.
\textcolor{black}{Even using the gradient accumulation strategy to enable a relatively feasible computational setting, the results are still unsatisfactory shown in Table.~\ref{tab:lengthLSTM}.
It may be caused by some inherent problems in LSTM training, for example, hidden state hardly well preserves long-range information.
Our long-range memory bank stores all the past frame information, and bypasses modeling long-range cues by LSTM.}
Cooperating with non-local bank operator, long-range motion cues from the memory bank can also be used without interrupting the end-to-end training.
Thanks to dispensable back-propagation for memory bank, \textcolor{black}{supportive feature} length reaches 40 seconds with 40 batch size, without extra computational cost involved.

Regarding the cholecystectomy procedure, one interesting phenomenon is shown by our results that keeping increasing the input length fails to continuously improve performance. 
\textcolor{black}{Around 30-second supportive features can largely increase recognition results to the best for the minute even hour-level cholecystectomy surgery.
There are two underlying reasons.
First, actions with quite fast motion form the majority in this surgery, such as using \textcolor{black}{hook} to dissert tissue. Therefore, 30-second supportive features can already cover one even multiple complete surgical motions for accurate recognition. As shown in Fig.~\ref{fig:intro} (b), 30-second frames can capture the complete motions in the fine-grained granularity levels (e.g., ‘action’ and ‘step’). 
Second, there exist lots of irrelevant noise from lens cleaning process and artefacts from smoke and blood in each phase.
Considering excessive long-range duration in cholecystectomy shall inevitably involve some artefacts when recognizing actions, leading to the performance degradation in the majority part and finally causing the overall performance with slight improvement.
However, we believe that further increasing the encoded temporal range is a promising direction to be explored for workflow recognition task. 
In the future, we plan to employ our general memory bank with longer information to other types of surgery which have relatively slow operation motions, to see whether it is due to surgery property. 
We also plan to completely replace LSTM by more powerful methods, e.g., Transformer, or combining truncated back propagation to further largely increase covered temporal span, which may further improve the recognition smoothness.}

Surgical workflows are typically well structured and ordered, according to the specified operation instructions.
\textcolor{black}{Although requiring the pre-defined prior information of surgical workflow order,}
incorporating such valuable domain knowledge can enhance the prediction consistency to boost performance.
Some previous works adopt simple yet effective post-processing strategies, such as averaging smoothing in~\cite{cadene2016m2cai} and prior knowledge inference, called PKI, in our previous work~\cite{jin2018sv,jin2020multi}.
\textcolor{black}{In this work, all the results reported above are purely predicted by a single TMRNet without any post-processing strategies.}
We have also conducted experiments to investigate the effect of post-processing on our method.
\textcolor{black}{We utilize PKI and peak the results of TMRNet on Cholec80 at 92.1\% PR, 91.1\% RE, and 93.7\% AC, surpassing the state-of-the-art refined results of 91.6\% PR, 90.1\% RE and 93.3\% AC in~\cite{jin2020multi}.}

Our long-range memory bank is pre-calculated over the whole complete surgical procedures and stored offline for network training.
Such external storage with a single computation pass avoids repeat and redundant cost for training.
To ensure the clinical usability of our model with real-time recognition, in inference, our memory bank is online updated by gradually archiving the feature of last frame.
The online update scheme has a negligible effect on recognition efficiency.
To this end, our whole model can recognize phase at a quick speed (around 0.08 s per frame with one GPU), which can be integrated into the real-time context-aware system and assist surgeons in real-world surgical operating.
In general, our proposed TMRNet not only can analyze the cholecystectomy, but also can be extended to address other types of surgical videos, such as cataract surgery, robotic surgery, etc.
Meanwhile, our method can be applied to many other situations, where long-range temporal information is crucial, such as microscopy cell tracking~\cite{ulman2017objective}, cardiac motion estimation~\cite{yu2020motion}, and ultrasound sequence analysis~\cite{yang2017fine}.
We believe our method can inspire more and further investigations on how to effectively analyze motions in the field of medical sequential data analysis.

\section{Conclusion}
We propose a novel temporal memory relation network (TMRNet) to recognize workflow from surgical videos, which wisely integrates long-range and multi-scale supportive information in temporal space for accurate recognition.
Specifically, our designed long-range memory bank provides long-term surgical cues.
Also, a temporal variation layer is designed to augment the long-range features from the memory bank to tolerate complex and multi-scale temporal extents.
Collaborated with the non-local bank operator, our TMRNet is capable of relating such rich supportive features to the current one, without disturbing the joint learning of spatio-temporal representations.
Extensive experiments on two typical surgical video datasets demonstrate the effectiveness of our method with a large performance gain, presenting the superiority over other state-of-the-art methods.

\bibliographystyle{IEEEtran}
\small\bibliography{refs}

\end{document}